\documentclass[10pt,twocolumn,letterpaper]{article}

\usepackage[pagenumbers]{cvpr} 

\usepackage{booktabs} 
\usepackage{enumitem}
\usepackage{makecell}
\usepackage{multirow}
\usepackage{dsfont}
\usepackage{cuted}
\usepackage{capt-of}
\usepackage{float}
\usepackage[hang,flushmargin]{footmisc}
\usepackage[accsupp]{axessibility} 
\usepackage[dvipsnames]{xcolor}

\renewcommand{\ddot}{\bar}

\newcommand{\bB}{\mathbf{B}}

\newcommand{\bd}{\mathbf{d}}

\newcommand{\bM}{\mathbf{M}}

\newcommand{\bp}{\mathbf{p}}

\newcommand{\bv}{\mathbf{v}}\newcommand{\bV}{\mathbf{V}}
\newcommand{\bW}{\mathbf{W}}
\newcommand{\bx}{\mathbf{x}}

\newcommand{\cG}{\mathcal{G}}

\newcommand{\cL}{\mathcal{L}}
\newcommand{\cM}{\mathcal{M}}

\newcommand{\cP}{\mathcal{P}}

\newcommand{\cV}{\mathcal{V}}

\newcommand{\figref}[1]{Fig.~\ref{#1}}
\newcommand{\secref}[1]{Section~\ref{#1}}
\newcommand{\eqnref}[1]{Eq.~\eqref{#1}}
\newcommand{\tabref}[1]{Table~\ref{#1}}

\makeatletter
\DeclareRobustCommand\onedot{\futurelet\@let@token\@onedot}
\def\@onedot{\ifx\@let@token.\else.\null\fi\xspace}

\makeatother

\definecolor{darkgreen}{rgb}{0,0.7,0}
\definecolor{darkblue}{RGB}{31,119,180}
\definecolor{darkred}{RGB}{214,39,40}

\newcommand{\anpei}[1]{\noindent{\color{black}{#1}}}

\definecolor{bronze}{rgb}{1,1,0.6}
\definecolor{silve}{rgb}{0.969,0.796,0.600}
\definecolor{gold}{rgb}{0.941,0.592,0.600}

\definecolor{scene}{rgb}{0.745,0.486,0.451}
\definecolor{probe}{rgb}{0.767,0.767,0.767}
\definecolor{core_v}{rgb}{0.791,0.876,0.723}
\definecolor{core_m}{rgb}{0.918,0.705,0.541}
\definecolor{basis_m}{rgb}{0.647,0.760,0.889}
\newcommand{\scene}[1]{\noindent{\color{scene}{\textbf{#1}}}}
\newcommand{\probe}[1]{\noindent{\color{probe}{\textbf{#1}}}}
\newcommand{\coreV}[1]{\noindent{\color{core_v}{\textbf{#1}}}}
\newcommand{\coreM}[1]{\noindent{\color{core_m}{\textbf{#1}}}}
\newcommand{\basisM}[1]{\noindent{\color{basis_m}{\textbf{#1}}}}

\newcommand{\boldstartspace}[1]{\vspace{0.1in}\noindent\textbf{#1}}

\definecolor{cvprblue}{rgb}{0.21,0.49,0.74}
\usepackage[pagebackref,breaklinks,colorlinks,citecolor=cvprblue]{hyperref}
\usepackage{amsmath}
\usepackage{afterpage}
\usepackage{float}

\def\paperID{1399} 
\def\confName{CVPR}
\def\confYear{2024}

\title{NeLF-Pro: Neural Light Field Probes for Multi-Scale Novel View Synthesis}

\author{Zinuo You\textsuperscript{1}\qquad
Andreas Geiger\textsuperscript{2}\qquad
Anpei Chen\textsuperscript{$\dag$,1,2} \\
\textsuperscript{1}ETH Z\"urich \qquad \textsuperscript{2}University of T\"ubingen, T\"ubingen AI Center 
\\ {\tt\small {zinyou@ethz.ch} \quad {\{a.geiger,anpei.chen\}@uni-tuebingen.de}}
}

\begin{document}
\maketitle
\footnotetext{\noindent $\dag$ Corresponding author. }


\begin{abstract}
We present NeLF-Pro, a novel representation to model and reconstruct light fields in diverse natural scenes that vary in extent and spatial granularity. In contrast to previous fast reconstruction methods that represent the 3D scene globally, we model the light field of a scene as a set of local light field feature probes, parameterized with position and multi-channel 2D feature maps. Our central idea is to bake the scene's light field into spatially varying learnable representations and to query point features by weighted blending of probes close to the camera - allowing for mipmap representation and rendering. We introduce a novel vector-matrix-matrix (VMM) factorization technique that effectively represents the light field feature probes as products of core factors (i.e., VM) shared among local feature probes, and a basis factor (i.e., M) - efficiently encoding internal relationships and patterns within the scene.
Experimentally, we demonstrate that NeLF-Pro significantly boosts the performance of feature grid-based representations, and achieves fast reconstruction with better rendering quality while maintaining compact modeling.
Project page: \href{https://sinoyou.github.io/nelf-pro/}{https://sinoyou.github.io/nelf-pro}. 
\end{abstract}    
\section{Introduction}
\label{sec:intro}
Efficiently representing and reconstructing 3D scenes that support high-fidelity novel view synthesis is critical for computer vision and graphics applications, including virtual and augmented reality, cinematography, robotics and autonomous navigation. Neural Radiance Fields (NeRF)~\cite{Mildenhall2020ECCV} have emerged as an effective representation to model real-world objects/scenes by incorporating density with view-dependent radiance.

Despite the high rendering quality, NeRF takes days to reconstruct an object or scene. The follow-up works demonstrated significant success in speeding up the training process by relaxing the input domain to optimizable feature grids~\cite{AlexYuandSaraFridovich-Keil2022CVPR, Sun2022CVPR, Chen2022ECCV, Mueller2022SIGGRAPH}. To fit an unbounded scene into bounded MLP and feature grids, scene contraction~\cite{Zhang2020ARXIVc,Barron2022CVPR,Wang2023CVPR,Mueller2022SIGGRAPH} is commonly used to warp the input domain within a ball or cube of fixed size. Despite their impressive performance, the basic representations they use, such as MLPs~\cite{Barron2022CVPR}, VMGrids~\cite{Chen2022ECCV}, and HashGrids~\cite{Mueller2022SIGGRAPH, Wang2023CVPR}, are constrained to small-scale scenes (e.g., mip-NeRF360 and Free dataset), because these representations rely on a bounded centralized space.
To represent large-scale scenes, a common strategy is to subdivide the target scene into cells, each endowed with individual representations~\cite{Tancik2022CVPR, Xiangli2022ECCV,Turki2022CVPR, Andreas2023CVPR,Gao2023ICCV}. However, determining the ideal position and dimensions of the local bounding boxes is not trivial. Employing bounding boxes with hard partitions can precipitate unstable optimization, giving rise to ``floater" artifacts as demonstrated in KiloNeRF~~\cite{Reiser2021ICCV}. To solve this, regularization terms such as geometrical pixel partition~\cite{Turki2022CVPR} or total variance loss~\cite{Xu2023CVPR} are usually needed to stabilize the reconstruction. On the other hand, soft partitions with overlaps require smoothing filters, such as appearance matching~\cite{Tancik2022CVPR} or per-frame blending~\cite{Andreas2023CVPR} for the colors predicted from different local representations, resulting in slow optimization.

\begin{figure}[!t]
    \centering
    \resizebox{\linewidth}{!}{\includegraphics{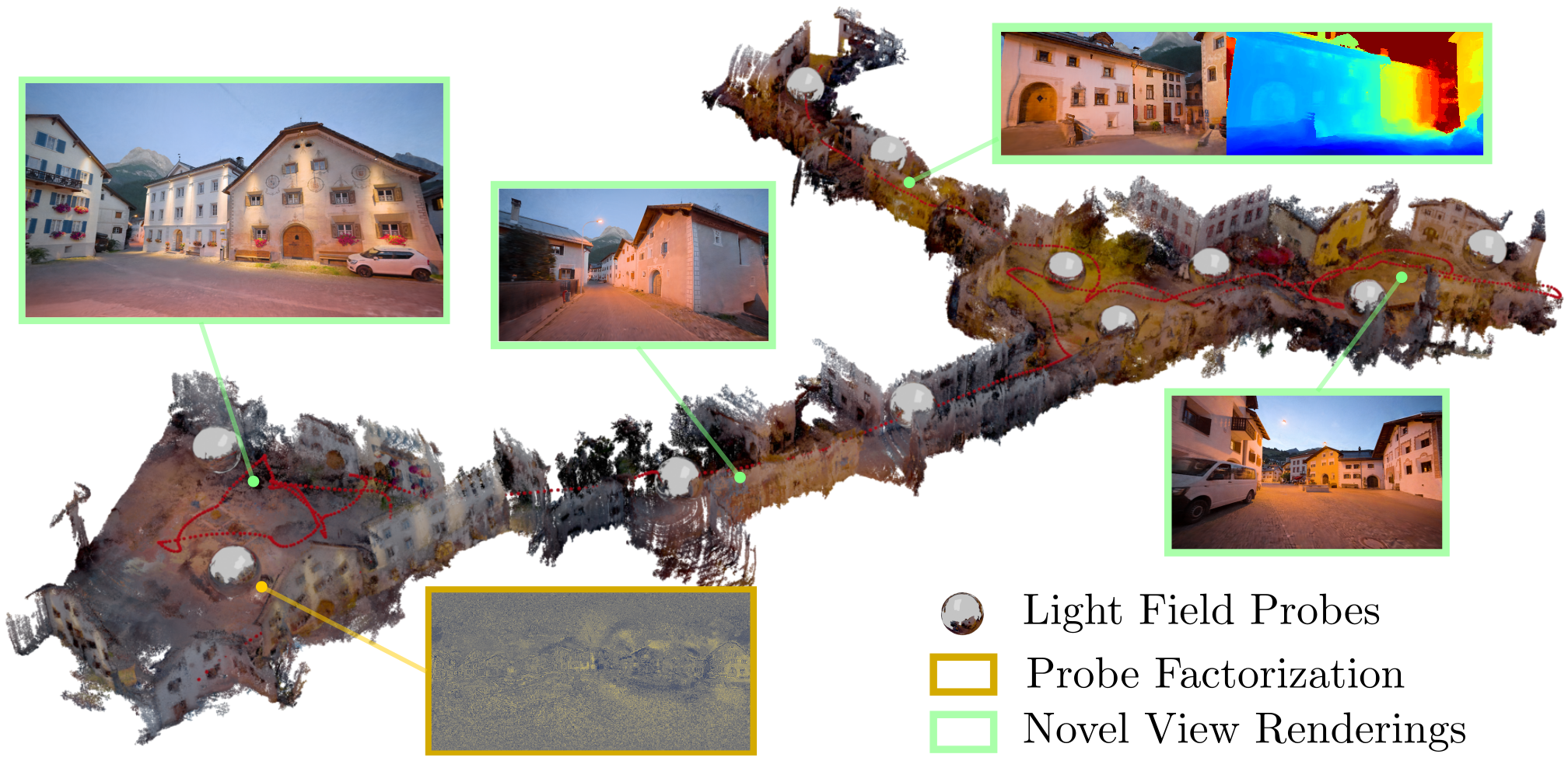}}
    \caption{\textbf{NeLF-Pro} represent a \scene{Scene} as spatially distributed \probe{Light Field Probes} for faithful novel view synthesis in diverse and spatially inhomogeneous natural scenes.}
    \label{fig:teaser}
\end{figure}

In this work, we propose \textbf{Ne}ural \textbf{L}ight \textbf{F}ield \textbf{Pro}bes (aka, \textbf{NeLF-Pro}), a novel representation that is compatible with small and large scale scenes, and at the same time achieves on-par-with or better-than current state-of-the-art novel view rendering quality. Unlike previous methods that encode positions globally into multilevel inputs, we decompose the scene into local probes, allowing for scale- and trajectory-agnostic modeling. The core idea of our representation is to jointly encode the light field and geometry using Light Field Probes and optimize the probes by gradient descent. Unlike standard light field methods that model rays using global parametrization, our light field probes capture visible rays in local areas, achieving occlusion-aware modeling and rendering.

To do so, we first decompose the global coordinates into local ones. 
We observe that encoding a large scene globally within a compact representation, such as VMGrid or HashGrid, necessitates a substantial number of factors or levels. Consequently, the entire representation must be loaded when querying features, leading to inefficiencies in computations and GPU memory utilization. To address these issues, we propose to decompose the scene into multiple local representations based on the camera trajectory. 
This leads to flexible scene modeling when dealing with diverse natural scenes that vary in extent and spatial granularity.
%
%


We leverage the fact that a full light field can be seen as an integration of light field samples, where each light field sample can be approximated by blending a collection of discrete light field images - which have been widely applied in Image Based Rendering (IBR)~\cite{Levoy1996SIGGRAPH}. We therefore present a novel light field feature probe representation to encode light field radiance and geometry as multiple-channel spherical feature maps. In this case, light field samples can be approximated by aggregating probe features. 
Unlike the recent popular scene contraction mechanism~\cite {Barron2022CVPR}, our work relaxes the assumption of a single scene center by retaining multiple representation centers. With this approach, we locally encode the radiance and geometry of the scene, thereby facilitating decomposition and reconstruction.

Our work aims to improve grid-based representations to handle scenes with arbitrary scales within a compact and efficient framework. Towards this goal, we present a novel \textbf{Vector-Matrix-Matrix} (VMM) decomposition technique that effectively represents the light field feature probes as the products of a core factor (i.e., VM) which is shared among local feature probes, and a basis factor (i.e., M) which is designated exclusively for each probe. The spatial sharing mechanism facilitates the uncovering of hidden relationships and patterns within the scene, while reducing the number of light field feature probes required to maintain expressivity.
With VMM factorization, our work compactly encodes the entire scene content within dispersed light field feature probes across the spatial domain, achieving compact encoding while also preserving locality.

We further propose a soft localization and blending algorithm to enable fast reconstruction and aliasing-free rendering. Specifically, we aim to construct a target image by selecting $K$ neighboring feature probes based on the distance between the target camera center and the probe centers. Then, instead of representing each probe feature individually through factor multiplication, we apply efficient weighted blending on each factor group before multiplication.  
%
It should be noted that our feature query differs in a principled way from previous grid-based representations. We index point features on camera-adjacent probes rather than point-adjacent probes. Together with the local spherical projections, this enables storing of a 3D point at different probes in a form of multi-scale pyramid (also known as mipmaps) - nearby areas are modeled at higher resolution, while distant areas are represented at lower resolution. This conforms to the perspective projection and allows querying the mip feature without tracing cones~\cite{Barron2022CVPR, Hu2023ICCV} or multiple samples~\cite{Barron2023ICCV}.
%
Additionally, the neighboring probe selection scheme demonstrates 3D scene locality and is able to facilitate training and rendering on a large scale without the need to load the entire representation into memory.
%
For continuous light field factorization, as inspired by TensoRF~\cite{Chen2022ECCV}, we apply linear factor interpolation for the query points.

We conduct extensive experiments on diverse scenes, including both small-scale, i.e., mip-NeRF360~\cite{Barron2022CVPR}, Free~\cite{Wang2023CVPR}, middle-scale KITTI360~\cite{Liao2022PAMI} dataset) and large-scale, i.e., Google Earth~\cite{Xiangli2022ECCV}, KITTI-360-large~\cite{Liao2022PAMI} and website video. We demonstrate that all models are able to achieve realistic novel view synthesis results that are on-par-with or better than state-of-the-art methods, especially on large-scale scenarios, providing noticeable improvements while preserving fast reconstruction and high compactness.

\section{Related Works}
\label{sec:realated}

\boldstartspace{Light fields.}
Light fields capture the spatial-angular distribution of radiance in a scene. They were first introduced by Levoy and Hanrahan and enable novel view synthesis without the need for explicit 3D models~\cite{Levoy1996SIGGRAPH}. Gortler et al.~\cite{Buehler2001SIGGRAPH,Wood2000SIGGRAPH,Coombe2005ESRT,gortler2023lumigraph} extended this idea with proxy geometry and improved the rendering quality using approximate geometry.
The advent of neural networks brought a paradigm shift in light field rendering and early works~\cite{Chen2018SIGGRAPH,Oechsle2020ARXIV} use MLPs to implicitly model light fields and directly predict ray colors. ~\cite{li2021neulf,wang2022r2l} propose novel light field representations for efficient rendering. Similarly, \cite{Zhou2018SIGGRAPH,Mildenhall2019SIGGRAPH} propose to predict blending weights based on image features and ray distances. 

Directly representing a full light field with a neural network is extremely difficult because of its complexity. To address this problem, Light Field Networks~\cite{Sitzmann2021NEURIPS} leverage an implicit MLP together with 6D Plücker parameterization of the space of light rays, allowing 360-degree light field modeling and demonstrating success in representing scenes and objects with simple texture. Subsequent works~\cite{Wang2021CVPRb,Kellnhofer2021CVPR,Suhail2022CVPR,Attal2022CVPR,Wizadwongsa2021CVPR} incorporate light field rendering with volumetric integration to achieve high fidelity rendering, especially in high reflectent and translucent regions. However, they are limited to forward-facing or small-scale synthetic scenes.

Light Field Probes were designed to encode a scene's full light field and internal visibility by placing a probe array in the scene, allowing a quick rendering of global illumination effects~\cite{Morgan2017I3D,Ari2017TOG}. Our work is inspired by probe representation. However, unlike traditional rendering setups, our work assumes that the geometry and radiance of the scene are unknown and attempts to learn and reconstruct the scene with gradient descent. We refer the readers to ~\cite{Ramamoorthi2023ARXIV,Tewari2022CGF} for a more comprehensive view.

\boldstartspace{Neural radiance field.}
Advanced neural representations are revolutionizing the reconstruction field as a promising replacement for traditional representations. They show significant progress in improving reconstruction quality and efficiency in a wide range of graphics and vision applications, including novel view synthesis~\cite{Zhou2018SIGGRAPH, Lombardi2019SIGGRAPH, Thies2019TOG, Aliev2019ARXIV, Mildenhall2020ECCV, Liu2020NEURIPS, Chen2022ECCV, Xu2022CVPRb,Verbin2022CVPR}, surface reconstruction~\cite{Niemeyer2019ARXIV, Yariv2021NEURIPS, Wang2021NEURIPSa, Yu2022NEURIPS,Yu2022SDFStudio}, appearance acquisition~\cite{Zhang2021CVPRb,Boss2021CVPR,Boss2021NEURIPS,Zhang2022CVPRb,Jin2023CVPR}, etc. In contrast to standard explicit representations such as meshes, point clouds, implicit representations \cite{Mescheder2019CVPR, Park2019CVPR, Chen2019CVPR} provide continuous and smooth representations of surfaces. 

NeRF \cite{Mildenhall2020ECCV} represents the neural radiance field for novel view synthesis using MLPs with positional encoding for scene representation. However, its reconstruction process is time-consuming. To address this, recent works have introduced learnable feature grids~\cite{AlexYuandSaraFridovich-Keil2022CVPR,Sun2022CVPR,Chen2022ECCV,Mueller2022SIGGRAPH} for faster optimization, achieving reconstructions in just 5 minutes. 
However, these representations, along with state-of-the-art methods~\cite {Barron2022CVPR,Barron2023ICCV}, assume object-centric scenarios and use scene contraction, which limits their scalability. 
To achieve compact reconstruction and constrain the reconstruction process, TensoRF uses various matrix factorizations to model a radiance field. However, it is scene-centric and fails to model large-scale scenes.
In contrast, F$^2$-NeRF~\cite{Wang2023CVPR} identifies a 3D subspace where the distance between any two points is equal to the sum of their 2D perspective distances across all visible cameras. However, F$^2$-NeRF still depends on the HashGrid representation, which limits its suitability for long trajectories or large-scale scenes. Our approach, on the other hand, defines the target space as a set of local 2D subspaces (light field probes) rather than a homogeneous 3D space. This naturally preserves the perspective distances and provides flexibility when modeling scenes with different scales.

\boldstartspace{Large-scale 3D reconstruction.}
Numerous works have succeeded in reconstructing large-scale scenes using standard structure-from-motion techniques or recent neural radiance fields. Early work~\cite{Agarwal2009ICCV} presents a system that can match and reconstruct entire cities from images collected from the Internet. While~\cite{Lafarge2012creating} improve accuracy, they still mainly focus on reconstructing point clouds with diffuse color, and use Poisson reconstruction to obtain meshes, lacking the ability to render realistic images from reconstructed representations.

Recent studies~\cite{Tancik2022CVPR,Xiangli2022ECCV,Turki2022CVPR,Mi2023ICLR,Andreas2023CVPR,Xu2023CVPR} have integrated NeRF with the concept of decomposing large-scale scenes into a union of local blocks while maintaining individual representations for each block. In practice, this localized approach introduces optimization challenges, often requiring the incorporation of smoothing regularization or additional supervision to enhance the stability.
To address challenges such as inconsistent exposure and moving objects in the capture of large scenes, NeRF-W~\cite{Martin-Brualla2020ARXIV} and Block-NeRF~\cite{Tancik2022CVPR} present an uncertainty and latent appearance modeling, which significantly reduce the reconstruction ``floaters".
It is worth mentioning that 3D Gaussian Splatting~\cite{Kerbl2023TOG} doesn't rely on scene contraction but with a projection mechanism, showing potential for large-scale scene reconstruction. However, the primitive-based representation can easily overfit to the training views and thus requires a good initial point cloud and dense training views. Our representation is between local radiance fields and primitive representations, utilizes only a few dozen local probes, and does not require a point cloud prior. 

\newcommand{\Point}{\bx}
\newcommand{\Center}{\bp_l}
\newcommand{\Dir}{\bd}
\newcommand{\LF}{\cL(\Point,\Dir)}
\newcommand{\DLF}{\cL_l(\Dir)}
\newcommand{\Probes}{\cL_l(\omega)}
\newcommand{\Weight}{\cw(\Point)}
\newcommand{\RayCast}{\Gamma}
\newcommand{\RRR}{\mathbb{R}}
\newcommand{\timenear}{t_n}
\newcommand{\timefar}{t_f}
\newcommand{\Transmit}{T}
\newcommand{\absrp}{\sigma}
\newcommand{\numsamples}{N}
\newcommand{\interval}{\Delta}
\newcommand{\expo}[1]{\exp\left(#1\right)}
\newcommand{\norm}[1]{\left\lVert#1\right\rVert}

\newcommand{\Mat}{\bB}

\newcommand{\Tensor}{\mathcal{T}}
\newcommand{\tensor}{\mathcal{\tau}}

\begin{figure*}[!t]
    \centering
    \includegraphics[width=0.93\linewidth]{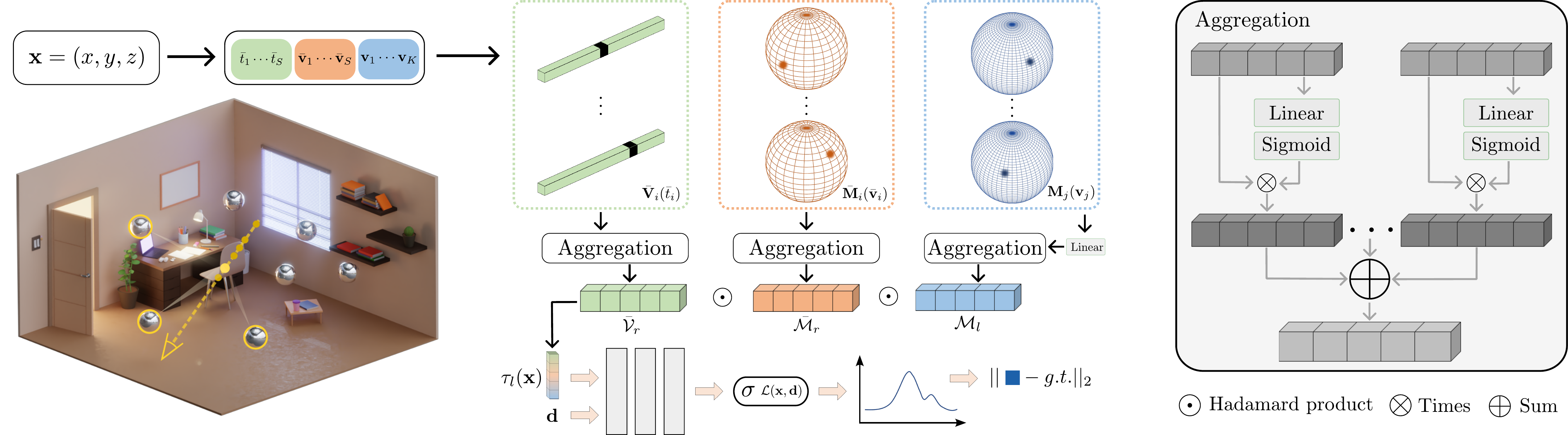}
    \caption{\textbf{NeLF-Pro pipeline.} Our approach models a scene using light field probes represented by a set of core vectors \coreV{$\ddot{\bV}$}, matrices \coreM{$\ddot{\bM}$} as well as a set of basis matrices \basisM{$\bM$}. We project the sample points onto local probes to obtain local spherical coordinates and then query factors from probes near the camera. These factors are aggregated using a self-attention mechanism and combined using Hadamard products. The resulting factor $\tensor_l(\bx)$ and the viewing direction $\bd$ are used to calculate the density $\sigma$ and the light field radiance $\LF$.} 
    \label{fig:pipeline}
\end{figure*}

\section{Neural Light Field Probes}

Our goal is to synthesize the novel views of the diverse scenes that vary in scale and spatial granularity, from a collection of posed multi-view images.
In the following sections, we first present our scene parameterization which bridges the gap between light field probes and volume rendering. We then introduce a factorization algorithm that efficiently factorizes radiance and geometric information for light field queries. Finally, we describe in detail how to distribute the probes during initialization and select them during querying for high-fidelity rendering.

\subsection{Volumetric Light Field Rendering}
\label{sec:parametrization}
A continuous light field $\LF$ encodes the spatial-angular distribution of radiance for all receivers $\Point \in \RRR^3$ over the unit outgoing directions $\Dir \in \mathbb{S}^2$. We build a sparse set of discrete representations of this distribution: $\cL_l(\Dir):= \cL(\bp_l,\Dir), l\in[1,\cdots, L]$ (\textit{Light Field Probes}), where each probe captures the angular variation as seen from the probe position $\Center$. The light field at any receiver $\Point$ can be reconstructed by calculating a spatial average from the radiance of the probes.

\begin{equation}
    \LF \approx \sum_{l=1}^L \mathrm{w}_l(\Dir) \cL_l(\Dir)
    \label{eqn:light_field}
\end{equation}
where $\mathrm{w}_l(\Dir) = \mathrm{\Tilde{w}}_l(\Dir)/\sum_k \mathrm{\Tilde{w}}_k(\Dir)$, which denotes the normalization weighting depending on the distance between $\bx$ and $\bp_l$.



Note that the main difference between light field and radiance field~\cite{Mildenhall2020ECCV} rendering is that, light field rendering directly utilizes the distribution of light in ray space, while radiance field reconstructs the light distribution in a volumetric space by integration.

Volumetric light field rendering combines the advantages of both methods and proposes to divide the target ray into a sequence of intervals between the near and far ranges and to sum over the light field segments. The light segment of each interval is determined by weighted blending of light field samples from the multi-view images:

\begin{gather}
 \LF 
 \approx \sum_{i=1}^{\numsamples} \Transmit_i \alpha_i \cL(\bx_i,\Dir)  
 \label{eqn:volume_rendering}\\
 \text{where} \quad \cL(\bx_i,\Dir) = \sum^L_{l=1} w_{i,l}\cL_l(\bv_{i,l})
\end{gather}
where $\Transmit_i$ are the approximated accumulated transmittances defined by $\Transmit_i=\prod_{j=1}^{i-1} (1-\alpha_j)$, and $\alpha_i = 1 - \expo{-\absrp_i\interval_i}$ is the approximated opacity value of sub-interval $\interval_i = t_{i+1} - t_{i}$, and $\absrp_i$ denotes the volume density of point sample $\bx_i = \bx + t_i \Dir$. $\bv_{i,l} = (\bx_i - \bp_l)/\norm{\bx_i - \bp_l}_2$ is the normalized direction between the sample point and the position of light field probes, $w_{i,l}$ are the normalized blending weight of the intervals. 



To obtain the density and blending weight at position $\bx_i$, a common solution is to reuse the input images as light field samples and to train a neural network $\Phi$ to predict the density and blended radiance from a set of light field queries ${\cL_l(\bx_i,\bv_{i,l})}$:

\begin{gather}
     \alpha_i, w_{i,l} = \Phi(\{\cL_l(\bv_{i,l})\})
    \label{eqn:network_weighting}
\end{gather}

However, these neural Image Based Rendering methods like IBRNet~\cite{Wang2021CVPRb} can easily lead to jumping artifacts when moving the camera and fail to accurately estimate geometry due to appearance bias.
Instead of directly using images, we introduce light probes to model light field samples. Our work assumes that the fusion network $\Phi$ and the light field probes $\cL_l(\bv)$ are both unknown and optimizes these parameters by gradient descent.
In practice, the light field probes and the fusion network can easily be overfitted to the multi-view inputs due to the strong flexibility and view-dependent nature of the light field. 




\subsection{Light Field Factorization}
\label{sec:decomposition}

We now present our novel light field probe representation for stable light field reconstruction. Our model consists of spatially distributed feature probes, each representing the radiance and geometry as multi-channel features.


We consider the discrete spherical probes as a set of 3D tensors $\Tensor = \{\tensor_l \in \RRR^{H \times W \times D}\}^L_{l=1}$, where each local probe $\tensor_l$ denotes a $D$-layered spherical grid located at $\bp_l$, $W, H$ are the spherical dimensions, providing full angular coverage. We propose to use low-dimensional factors to represent the light field tensors $\Tensor$ for compact modeling, and introduce a novel vector-matrix-matrix (VMM) factorization, where a factor $\ddot{V},\ddot{M}$, serves as core factors and captures the multilinear interactions between different factors, accompanied by basis factors $M$ located at position $\bx$. VMM factorization factorizes each tensor element $\tensor_l^{ijk}$ as a sum of $R$ scalar products:

\begin{equation}
 \tensor_l^{ijk}
 = \sum^R_{r=1}\ddot{\bV}_{c,r}^{k} \circ \ddot{\bM}_{c,r}^{ij} \circ \bM_{l,r}^{ij}
 \label{eqn:light_field_factorization}
\end{equation}
where the factor $\ddot{\bV}_{c,r} \in \RRR^{D}$ is a vector of length $D$, and the matrix $\ddot{\bM}_{c,r} \in \RRR^{H \times W}, \bM_{l,r} \in \RRR^{H \times W}$ are matrix tensors, $\circ$ denotes the Hadamard product. We use $c,l$ as the index indices core and basis factors respectively, $i,j,k$ denote the indices of the three modes.

\boldstartspace{Distributing factors} 
We distribute both the core and basis factors as scene representation in 3D, and apply the Farthest Point Sampling~(FPS)~\cite{Eldar1997TIP} algorithm to select $L$ positions for the basis factor from the input cameras, and then use the K-means algorithm to select $C$ positions for the core factors. The number of factors $C,L$ can be set differently and should be chosen depending on the complexity of the scene. Inspired by the factor sharing mechanism in Tucker Decomposition~\cite{Tucker1966Springer}, we propose to share the core factor across different basis factors, resulting in $C \ll L$.
The sharing mechanism enables the discovery of hidden relationships and patterns within the scene.

Unlike previous scene representations, which are usually centered at the scene, our approach exploits the prior information provided by the input camera positions, facilitating the handling of scenes with arbitrary shapes. 
By distributing local scene representations in these regions, our method enhances representation efficiency. 
%

\boldstartspace{Continuous factorization.}
\label{sec:continue_factorization}
To achieve a continuous field, we employ linear interpolation on each vector and matrix while querying the tensor grid features of a target position $\bx$, combining with ~\eqnref{eqn:light_field_factorization} we obtain, 
\begin{equation}
 \tensor_l(\bx)
 = \sum^R_{r=1} \ddot{\bV}_{c,r}(\bx) \circ \ddot{\bM}_{c,r}(\bx) \circ \bM_{l,r}(\bx)
 \label{eqn:continue_factorization}
\end{equation}
where $(\cdot)$ denotes linear interpolation.

\boldstartspace{Coordinate transformation.}
\label{sec:transformation}
To query values on the low-dimensional factors, we transform the input coordinates $\bx$ to local coordinates using a projection transformation. For a 3D location denoted by $\bx=(x,y,z)$, we compute its distance to a factor center $\Tilde{t} = \norm{\bx - \bp}_2$ and transform it into local normalized depth and spherical coordinates as follows:

\begin{equation}
 t = \frac{1}{\Tilde{t}+1}, 
 \Tilde{\bv} = \left(\arccos\left(\frac{z}{\Tilde{t}}\right),\arctan\left(\frac{y}{x}\right)\right)
 \label{eqn:transformation}
\end{equation}
Following Factor Fields~\cite{Chen2023factor,Chen2023SIGGRAPH}, we apply a scaling factor and a periodic function to the projected coordinate to effectively utilize the empty regions within the spherical grids, $\bv = \gamma(\nu \cdot \Tilde{\bv})$, where $\nu$ determines the frequency and $\gamma$ is the sawtooth function.


Therefore, the tensor values at location $\bx$ denoted in ~\eqnref{eqn:continue_factorization} are given by:
\begin{equation}
 \tensor_l(\bx)
 = \sum^R_{r=1} \ddot{\bV}_{c,r}(\ddot{t}_{c}) \circ \ddot{\bM}_{c,r}(\ddot{\bv}_c) \circ \bM_{l,r}(\bv_{l})
 \label{eqn:feature_intepolation}
\end{equation}
where $\ddot{t}$ and $\ddot{\bv}$ represent the projected coordinates of the core factors.

To simplify~\eqnref{eqn:feature_intepolation}, we concatenate all scalar values into an $R$-channel vector, this yields $\ddot{\bV}_{c}(\ddot{t}_{c}) = [\ddot{\bV}_{c,1}(\ddot{t}_{c}),\cdots,\ddot{\bV}_{c,R}(\ddot{t}_{c})]$. Similarly, we define $\ddot{\bM}_{c}(\ddot{\bv}_c)$ and $\bM_l(\bv_{l})$. In the following, we denote $\cG_l(\bx) = \ddot{\bV}_{c}(\ddot{t}_{c}) \circ \ddot{\bM}_{c}(\ddot{\bv}_c) \circ \bM_l(\bv_{l})$. If necessary, we can recover $\tensor_l(\bx)$ from $\cG_l(\bx)$ using a simple sum operation $\tensor_l(\bx) = \sum \cG_l(\bx)$, though this operation is omitted with the inclusion of a neural decoder. 

\boldstartspace{Localization.}
We now detail how to form an image using a subset of probes, resulting in low memory consumption and scalable representation. 
When the scene size and consequently the number of necessary probes increases, rendering becomes inefficient.
%
 
Therefore, we propose to select the $\hat{C}$ nearest core factors and the $\hat{L}$ basis factors out of $C$ and $L$ based on the distance between $\bp_l$ and the position of the target camera.
%
After that, only the selected factors are used when rendering a target view, enabling reconstruction and rendering of large scenes or using computationally restricted hardware.

In addition, such a strategy implicitly mimics a mipmap representation. Due to the spatially distributed light field probes and spherical projection transformation, the light field features of a 3D point are captured at multiple resolutions across different probes, where the \anpei{representation resolution is automatically adjusted based on the relative distance between the 3D point and the probe's center}. Rendering with the feature in neighboring probes naturally encodes a proper level of scene features, where 3D content close to the target view is represented at high resolution due to the nature of the spherical projection, as described in \secref{sec:transformation}.

\boldstartspace{Factor aggregation.}
\label{sec:feature_agg}
Our work leverages local modeling and regionally selects factors for target view rendering. Naively concatenating the selected factors can introduce a permutation-variant problem.
To address this problem, we propose a permutation-invariant factor blending scheme. Given the $\hat{C}$ selected shared core and $\hat{L}$ basis factors for each query point, we predict a weight for each factor and blend the factors of same groups using a self-attention mechanism:
%


\begin{equation}
\cG(\bx)
  = \ddot{\cV} \circ  \ddot{\cM} \circ \cM,\,\text{where} \left\{
  \footnotesize 
  \begin{aligned}
     \ddot{\cV} &= \sum_{i=1}^{\hat{C}} w_i^{\ddot{\bV}}\ddot{\bV}_{i}(\ddot{t}_i)\\
      \ddot{\cM} &= \sum_{i=1}^{\hat{C}} w_i^{\ddot{\bM}}\ddot{\bM}_{i}(\ddot{\bv}_i)\\
      \cM &= \sum_{j=1}^{\hat{L}} w_j^{\bM}\bM_{j}(\bv_j)
  \end{aligned}
  \right.
 \label{eqn:vm_fusion}
\end{equation}
where the $\cG(\bx) \in \RRR^R$ is the aggregated vector. 



we obtain the scaler blending weight $w$ by a trainable linear transformation with sigmoid activation,

\begin{equation}
w = \text{Sigmoid}\left(\bW f + b \right)
\label{eqn:tensor_blending_weight}
\end{equation}
here, $f = \ddot{\bV}_i(\ddot{t}_i), \ddot{\bM}_i(\ddot{\bv}_i)$, and $\bM_{j}(\bv_{j})$ respectively. 

To further boost the compactness, we reduce the component dimension of the basis factor from $R$ to $F$ by applying a linear layer to $\mathbf{M}_j(\mathbf{v}_j)$ before factor aggregation, as shown in~\figref{fig:pipeline}.




\boldstartspace{Decoder.}
So far, we have introduced the factorization of the light field features. However, unlike the direct light field rendering in Eqn.\eqref{eqn:volume_rendering}, which works with radiance samples, our light field representation is embedded in feature space. To obtain multi-dimensional outputs, we introduce a MLP network $\cP$ to obtain the radiance and geometric density:
\begin{equation}
 \alpha_i, \cL(\bx_i,\Dir)
 = \cP(\cG(\bx_i),\Dir)
 \label{eqn:projection_network}
\end{equation}

\subsection{Optimization}
\label{sec:optimization}
We optimize the light field feature probes and the projection network using gradient descent, and minimize a $L_2$ loss between the reconstructed images and the multiview inputs. Using global and local factorization as basis factors that correlate and regularize the entire field during optimization, our method achieves efficient reconstruction. Furthermore, we include the interlevel loss from mip-NeRF 360~\cite{Barron2022CVPR} in our optimization process.


\section{Experiments}
\subsection{Experimental Settings}
\begin{figure}
\centering
\captionsetup{justification=centering}

\begin{subfigure}[c]{0.40\linewidth}
\includegraphics[width=\linewidth]{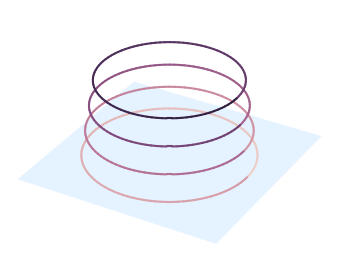}

\caption{small 360 degree circling \\ mip-NeRF360~\cite{Barron2022CVPR}}
\end{subfigure}
\hfill 
\begin{subfigure}[c]{0.40\linewidth}
\includegraphics[width=\linewidth]{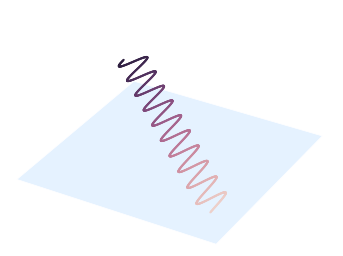}
\caption{short forwarding\\ Free~\cite{Wang2023CVPR} and KITTI360~\cite{Liao2022PAMI}}
\end{subfigure}
\begin{subfigure}[c]{0.41\linewidth}
\includegraphics[width=\linewidth]{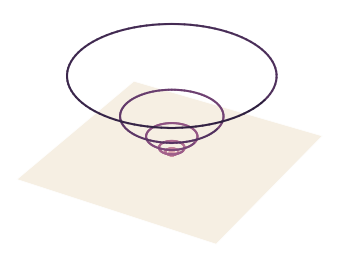}
\caption{aerial multi-scale circling\\ Google Earth~\cite{Xiangli2022ECCV}}
\end{subfigure}
\hfill
\begin{subfigure}[c]{0.41\linewidth}
\includegraphics[width=\linewidth]{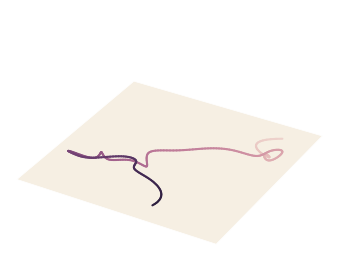}
\caption{free forwarding\\ KITTI360-big~\cite{Liao2022PAMI}, City Walk}
\end{subfigure}
\vspace{-2mm}
\caption{\textbf{Datasets and trajectories used for our evaluation.}}
\label{fig:trajectoies}
\end{figure}

\boldstartspace{Datasets}
We evaluate our method on four datasets spanning diverse scene scales and camera trajectories, including the mip-NeRF360 dataset~\cite{Barron2022CVPR}, the Free dataset~\cite{Wang2023CVPR}, BungeeNeRF~\cite{Xiangli2022ECCV}, and KITTI-360~\cite{Liao2022PAMI}, as well as a city walk video from \href{https://www.youtube.com/watch?v=kpXiVSmvbKg&t=10s}{Youtube}, their trajectories are shown in \figref{fig:trajectoies}. For the KITTI-360 dataset, we use two different scales: medium and large scales. For medium-scale scenes, we used five static sequences from Novel View Appearance Synthesis (50\% Drop), with a distance of about 50 meters, labeled as KITTI-360. In addition, we selected a long trajectory with a driving distance of approximately 1500 meters for a large-scale evaluation, labeled as KITTI-360-big.

\setlength{\tabcolsep}{5pt}
\begin{table*}[t]
\centering
\resizebox{0.95\textwidth}{!}{%

\begin{tabular}{lcc|cc|ccc|ccc}
&&& \multicolumn{5}{c|}{Free Dataset} & \multicolumn{3}{c}{mip-NeRF360 Dataset} \\
Method & Iterations & Batch Size & Time $\downarrow$  & \#Params $\downarrow$ & PSNR$\uparrow$ & SSIM$\uparrow$ &LPIPS $\downarrow$ & PSNR$\uparrow$ & SSIM$\uparrow$ & LPIPS $\downarrow$ \\
\hline

    NeRF++~\cite{Zhang2020ARXIVc}         & 250K & 2K & hours     & -      & 23.47 & 0.603 & 0.499 & 26.21 & 0.729 & 0.348 \\
    mip-NeRF360~\cite{Barron2022CVPR}        & 250K & 16K & hours     & 8.6M      & \textbf{27.01} & \textbf{0.766} & \textbf{0.295} & \textbf{28.94} & \textbf{0.837} & \textbf{0.208} \\
    \hline
    Plenoxels~\cite{AlexYuandSaraFridovich-Keil2022CVPR}     & 128K & 5K &25m       & Dynamic     & 19.13 & 0.507 &  0.543 & 23.35 &  0.651 & 0.471 \\
    DVGO~\cite{Sun2022CVPR}          & 30K & 5K & 21m       & 153M    & 23.90 & 0.651 &  0.455 & 25.42 &  0.695 & 0.429 \\
    Instant-NGP~\cite{Mueller2022SIGGRAPH}      & 20K & 10-85K & 6m        & 12M   & 24.43 & 0.677 &  0.413 & 26.24 &  0.716 & 0.404 
    \\
    F$^2$-NeRF~\cite{Wang2023CVPR}         & 20K & 10-85K & 12m       & 18M   & 26.32 & 0.779 &  0.276 & 26.39 &  0.746 & 0.361\\
    NeLF-Pro (ours)                                    & 20K & 4K & 16m     & 20M      & \textbf{27.02} & \textbf{0.802} & \textbf{0.256} & \textbf{27.27} & \textbf{0.753} & \textbf{0.360} \\
    
\end{tabular}
} 

\vspace{-3mm}

\caption{\textbf{Results on the Free and mip-NeRF360 dataset.} The scores of the baseline methods are taken from F$^2$-NeRF~\cite{Wang2023CVPR}. Our training times are evaluated on an RTX3090 GPU. }

\label{tab:small-scale-main-result}  

\end{table*}


%

%
\boldstartspace{Implementations}
We implement our NeLF-Pro with PyTorch within the SDFStudio framework~\cite{Yu2022SDFStudio,nerfstudio}, without customized CUDA kernels. We train our model on a single RTX 3090 GPU with Adam optimizer~\cite{Kingma2015ICLR} and a learning annealed in multiple steps from $10^{-2}$ to $3.5\times10^{-4}$. 
We use a two-layer proposal model~\cite{Barron2022CVPR} for sampling points along the rays, with $[256, 96, 48]$ samples per stage. The proposal model is represented with our sparse core factor and is optimized by a distillation loss~\cite{Barron2022CVPR}.

We implement the core and basis factors of light field probes with feature grids.
We set the resolution of the core vector $\ddot{\bV}$ to $64$ and the matrix $\ddot{\bM}$ to $(64, 128)$, progressively upscaling them to $1024$ and $(128, 256)$, respectively. For the basis factor, we opt for a lower resolution, fixed at $(256, 512)$ during the optimization process. For all experiments, we maintain the factor component number at $R=32$ and $F=2$.


Due to the difference in scale, the probe number $L$ and the core tensor number $C$ vary for different datasets. For small scale scenes, we set $(L, C)=(64, 3)$;$(100, 5)$;$(256,16)$ for the small, medium, and large scale scenario respectively. 
We use $\hat{L}=16$ neighboring basis factors to represent the target views. The core factor for neighbor selection $\hat{C}$ is 3 for centralized camera trajectories (mip-NeRF360 and 56Leonard) and 2 for others. The periodic factor $\nu$ is $(5,5)$ for the mip-NeRF360 dataset, $(5, 2)$ for KITTI middle scale and $(4,4)$ for others.

\begin{table}[t]

\resizebox{0.95\linewidth}{!}{

\begin{tabular}{l|ccc|c}
Method & PSNR$\uparrow$ & SSIM$\uparrow$ &LPIPS $\downarrow$ & Time $\downarrow$ \\
\hline

    NeRF~\cite{Mildenhall2020ECCV}         & 21.28 & 0.779 & 0.343 & days \\
    mip-NeRF~\cite{Barron2021ICCV}        & 21.54 & 0.778 & 0.365 & days  \\
    PNF~\cite{KunduCVPR2022CVPR}     & 22.07 & 0.820 &  0.221 & days  \\
    NeLF-Pro (ours)                                    & \textbf{22.55} & \textbf{0.829} & \textbf{0.201} & \textbf{$\mathbf{45}$mins} \\
    
\end{tabular}

}
\vspace{-3mm}
\caption{\textbf{Quantitative result on KITTI360-NVS-appearance-50\% task~\cite{Liao2022PAMI}.} We report SSIM and LPIPS from the leaderboard. }
\label{tab:kitti-result}

\end{table}

\subsection{Comparisons}
\boldstartspace{Small Scale Scenes.}
We consider the mip-NeRF360 and Free datasets as small-scale datasets and compare our method with MLP and primitive-based representations, partly with voxel-grid methods, as shown in~\tabref{tab:small-scale-main-result}. 

Our NeLF-Pro significantly minimizes the known performance gap between feature grid-based representations and the well-trained MLP SOTA on the unbounded scenes, achieving SOTA performance over feature grid-based methods~\cite{Mueller2022SIGGRAPH, Wang2023CVPR} on both the Free and mip-NeRF360 datasets, while it is on-par-with mip-NeRF360~\cite{Barron2022CVPR} on the Free dataset.

\begin{figure}[!h]
    \centering
    \resizebox{\linewidth}{!}{
        \includegraphics{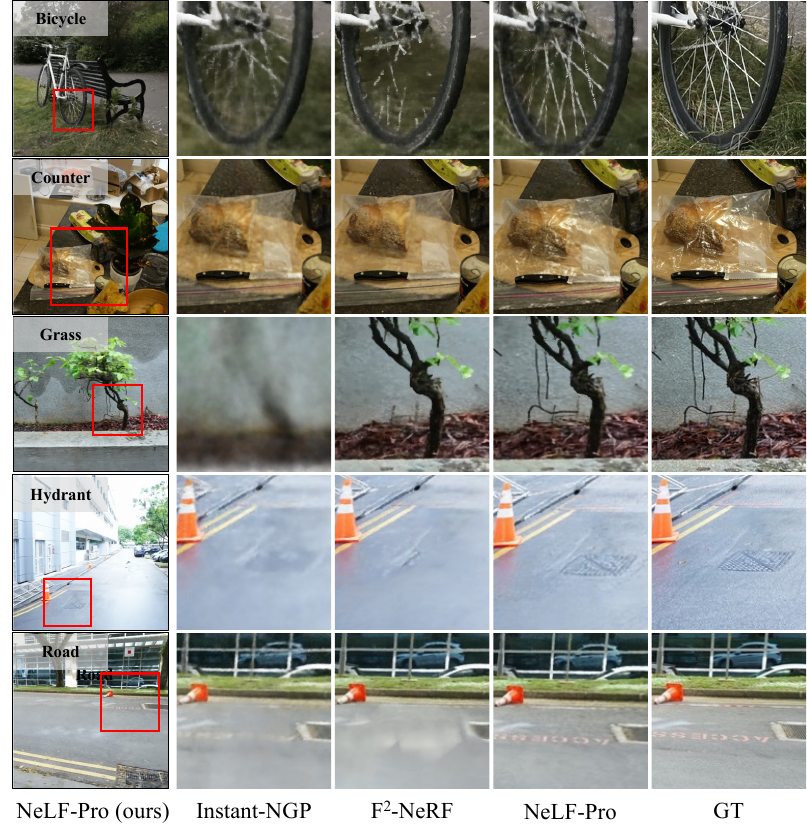}
    }
    \vspace{-9mm}
    \caption{\textbf{Qualitative results on mip-NeRF360~\cite{Barron2022CVPR} and Free dataset~\cite{Wang2023CVPR}.} Our approach is able to better reconstruct thin structure and appearance details than the baseline methods (Instant-NGP~\cite{Mueller2022SIGGRAPH}, F$^2$-NeRF~\cite{Wang2023CVPR}).}
    \label{fig:small-scale-qualitative}
\end{figure}

The qualitative results are shown in~\figref{fig:small-scale-qualitative}, our approach is capable of providing better renderings, especially in recovering thin structures. 
Thanks to the local representation, our method models specular reflections more accurately. Furthermore, our method is efficient and shows a training time comparable to Instant-NGP~\cite{Mueller2022SIGGRAPH} and F$^2$-NeRF~\cite{Wang2023CVPR}, both implemented with customized CUDA kernels. 
Note that F$^2$-NeRF requires precomputing near-far values for each training view from the calibration point cloud, while we instead use constant near-far values without the need for a sparse point cloud.

\begin{figure}
    \centering
    \resizebox{\linewidth}{!}{
        \includegraphics{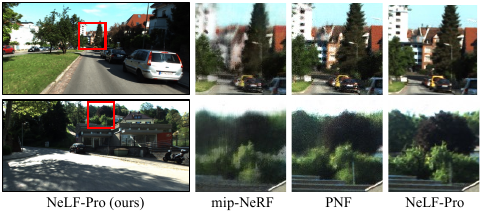}
    }
    \vspace{-9mm}
    \caption{\textbf{Results on KITTI-360 NVS Task.} Our approach provides more sharp renderings than the baselines.}
    \label{fig:kitti-qualitative}
\end{figure}

\boldstartspace{Medium Scale Scenes.}
Since the training views are accessible for the KITTI 360 dataset, we submitted our results to the leaderboard to obtain evaluation scores. In ~\tabref{tab:kitti-result}, we compare our method with other submissions that used the same setting as ours, i.e., RGB supervision only. Note that, in compliance with KITTI-360's guidelines, we couldn't submit results for other methods. Thus, our comparison is limited to previous submissions.
Our approach consistently outperforms all baseline methods for all three metrics and also provides more than 20 times faster training speed. 
~\figref{fig:kitti-qualitative} demonstrates that our approach yields sharper renderings compared to the baseline methods. 

\setlength{\tabcolsep}{5pt}
\begin{table*}[t]
\centering
\resizebox{0.95\textwidth}{!}{%
\begin{tabular}{l|cc|ccc|ccc|ccc}
&&&\multicolumn{3}{c|}{56-Leonard}  & \multicolumn{3}{c|}{Scuol}& \multicolumn{3}{c}{KITTI-360-big}\\
Method  &  Time $\downarrow$  & \#Params $\downarrow$ & PSNR$\uparrow$ & SSIM$\uparrow$ & LPIPS $\downarrow$ & PSNR$\uparrow$ & SSIM$\uparrow$ & LPIPS $\downarrow$ & PSNR$\uparrow$ & SSIM$\uparrow$ & LPIPS $\downarrow$\\
\hline
mip-NeRF360~\cite{Barron2022CVPR}  & 13.3h & 8.6M & 26.22 & 0.829 & 0.200 & 25.20 & 0.800 & 0.381 & 16.63 & 0.537 & 0.527\\
GridNeRF~\cite{Xu2023CVPR} & 6.9h & 276M & 23.42 & 0.731 & 0.318 & 23.65 & 0.763 & 0.510 & 18.55 & 0.548 & 0.607\\
3DGS~\cite{Kerbl2023TOG} & 2.5h & 225M & \textbf{29.23} & \textbf{0.942} & \textbf{0.075} & 22.74 & 0.835 & \textbf{0.301} & 15.17 & 0.510 & 0.601 \\
F$^2$-NeRF-Big~\cite{Wang2023CVPR} & 1h & 148M  & 27.07 & \underline{0.930} & \underline{0.102} & \underline{26.43} & \underline{0.842} & 0.385 & \underline{21.36} & \underline{0.686} & \underline{0.446} \\
NeLF-Pro (ours)  & 1.6h & 100M & \underline{28.86} & \underline{0.930} & 0.108 & \textbf{27.84} & \textbf{0.855} & \underline{0.368} & \textbf{22.68} & \textbf{0.720} & \textbf{0.399}\\
\end{tabular}
}
\vspace{-3mm}
\caption{\textbf{We show averaged results for large-scale scenes.} \textbf{Bold} score for the first and \underline{underlined} score for the second. Time and parameter number are the mean across all the scenes. In general, our method is on-par-with or outperforms other reconstruction methods. While 3DGS and F$^2$-NeRF-Big achieve similar or better results than ours on the 56-Leonard scene, they perform better near the center of the scene and less effectively for zoomed-out views. In contrast, our approach maintains high rendering quality throughout the entire trajectory due to its distributed representation. Our representation is also more compact and requires fewer parameters. }
\label{tab:large-scale-result}
\end{table*}

\begin{figure*}
\centering
\includegraphics[width=0.94\linewidth]{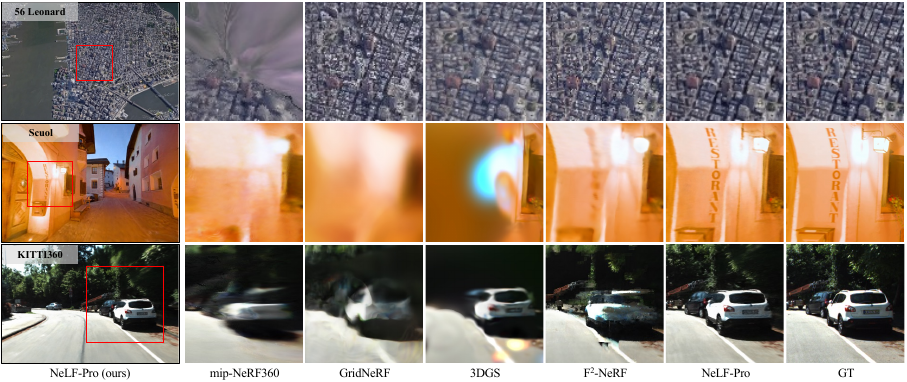}
\vspace{-3mm}
\caption{\textbf{Results on large-scale scenes.} Our representation better preserves fine structural details.}
\label{fig:large-scale-qualitative}
\end{figure*}

\boldstartspace{Large Scale Scenes.}
We also show the results of large-scale reconstruction for three distinctly different types of scenes: 56-Leonard with an input level of detail that includes a bird-eye-view, Scuol with a town in which randomly walking trajectory, and KITTI-360-big with a long straight driving style.
In our experiments, we use a four-fold larger model with three-fold more iterations (60k iterations) to reconstruct large scenes. We applied a similar increase in model size to the comparison methods.

The quantitative~\tabref{tab:large-scale-result} and visual results~\figref{fig:large-scale-qualitative} clearly demonstrate the high efficiency and strong scalability of our approach. Impressively, we found that our approach is able to better preserve structured patterns, which benefit from local representation and modeling of core factors.

\subsection{Ablation Study}
In \tabref{tab:ablation-study} we evaluate several ablations of NeLF-Pro: a-c) partial or complete removal of core factors $\ddot{V}$ modeling; d) ``point neighborhood" queries the nearest neighbor factors with the distance between the sample point and the center of probes, instead of the proposed camera neighborhood; e) ``random centers" distributes the light field feature probes by randomly selecting positions from the centers of the input cameras; f) ``w/o coordinate scaling $\nu$" disables the scali   ng of the spherical coordinates by setting the scaling factor $\nu=(1,1)$; g) ``factor concatenation" concatenates the selected factors before the products.

We observe that disabling the core factor $\ddot{V}$ results in a tiny loss of quality in small scene, while disabling core factor $\ddot{M}$, or the simultaneous removal of both factors leads to a significant drop in performance.
Moreover, selecting neighboring factors with the target camera outperforms the ``point neighborhood" strategy in the scene with multi-level images, benefiting from our mipmap-like representation.
Our factor aggregation model delivers significant performance improvements over the factor concatenation baseline.

\begin{table}[t]

\resizebox{\linewidth}{!}
{%
\begin{tabular}{ll|ccc|ccc}
&& \multicolumn{3}{c|}{Room} & \multicolumn{3}{c}{Scuol}
\\
&& PSNR$\uparrow$ & SSIM$\uparrow$ &LPIPS $\downarrow$ & PSNR$\uparrow$ & SSIM$\uparrow$ &LPIPS $\downarrow$ \\
\hline
& full model & 30.20 & 0.871 & 0.367 & 27.84 & 0.855 & 0.368 \\
a) & w/o core factor $\ddot{\bV}$ & 29.74 & 0.864 & 0.382 & 27.59 & 0.846 & 0.382\\
b) &w/o core factor $\ddot{\bM}$ & 28.16 & 0.818 & 0.434 & 24.55 & 0.741 & 0.518\\
c) &w/o core factors $\ddot{\bV},\ddot{\bM}$ & 23.78 & 0.655 & 0.550 & 22.76 & 0.639 & 0.595\\
d) &point neighborhood & 28.97 & 0.845 & 0.414 & 26.72 & 0.824 & 0.417\\
e) &random centers & 30.24&  0.871 &  0.365 & 27.81 &  0.853 & 0.371 \\
f) &w/o coordinate scaling $\nu$ & 28.65 & 0.830 & 0.465 & 26.97 & 0.827 & 0.421 \\
g) &factor concatenation & 27.68 & 0.764 & 0.477 & 24.78 & 0.694 & 0.547 \\
\end{tabular}
}
\vspace{-3mm}
\caption{\textbf{Ablation study} on a 360-degree surrounding room scene and the Scuol scene with free camera trajectories. }
\label{tab:ablation-study}

\end{table}
\section{Conclusion}
We presented a novel volumetric light field probe representation for high-quality scene reconstruction and rendering. We have also proposed a VMM factorization technique that decomposes light field probes into products of local factors, resulting in a compact representation with novel view synthesis on par with or better than state-of-the-art methods in diverse scenarios. 



\section{Acknowledgement}
Andreas Geiger and Anpei Chen are supported by the ERC Starting Grant LEGO-3D (850533) and DFG EXC number 2064/1 - project number 390727645. We thank Zehao Yu and Binbin Huang for proofreading and their insightful suggestions. We also thank Siyu Tang and Kaifeng Zhao for their valuable discussions and help. Special thanks to \href{https://www.youtube.com/@TheFlyingDutchMan8K}{The Flying Dutchman} for the city walk video. 
\clearpage
{
    \bibliographystyle{ieeenat_fullname}
    \bibliography{bibliography_short,bibliography,bibliography_custom}
}
\clearpage

\clearpage
\setcounter{page}{1}
\maketitlesupplementary
\appendix

In this \textbf{supplementary document}, we present visualizations of probe distribution, provide detailed analyses of the ablations along with observations, discuss the limitations and future work, and offer more quantitative and qualitative results.

\section{Light Field Probes Distribution.}
\begin{figure*}[!htp]
\centering

\begin{subfigure}[c]{0.3\linewidth}
\includegraphics[width=\linewidth]{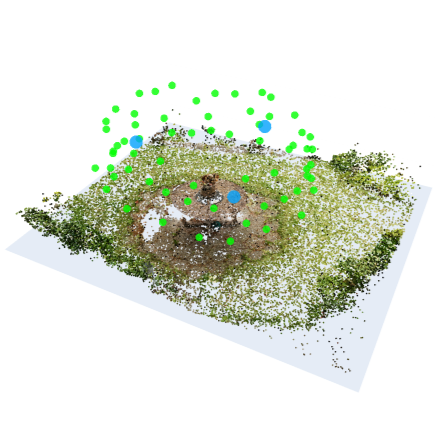}
\caption{\textbf{Garden}}
\end{subfigure}
\begin{subfigure}[c]{0.3\linewidth}
\includegraphics[width=\linewidth]{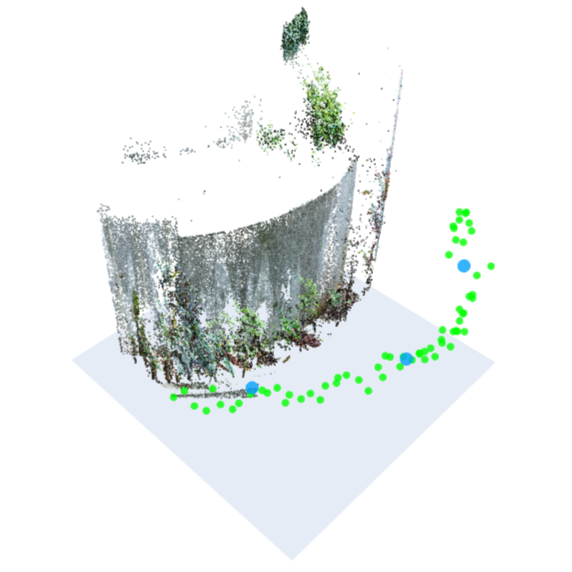}
\caption{\textbf{Grass}}
\end{subfigure}
\begin{subfigure}[c]{0.3\linewidth}
\includegraphics[width=\linewidth]{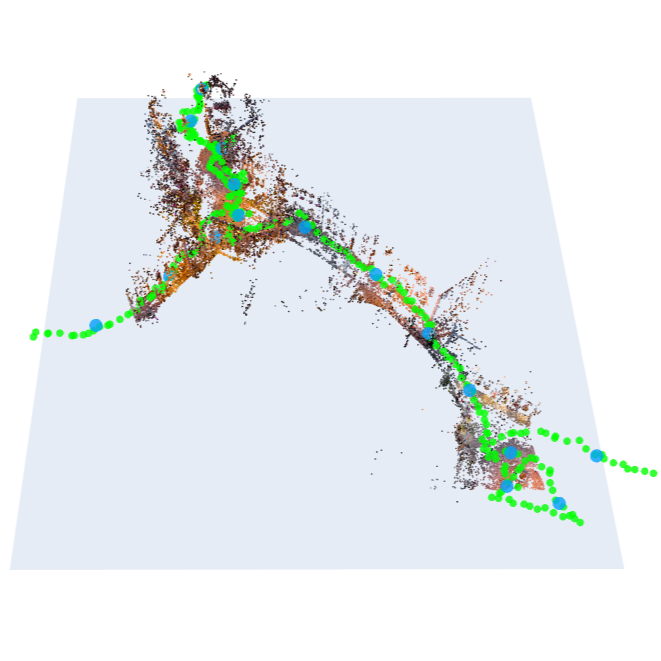}
\caption{\textbf{Scuol}}
\end{subfigure}

\vspace{-2mm}
\caption{\textbf{The distribution of light field feature probes.} The green spheres denote basis factors and the blue dots are core factors. We densely distribute the basis factors in the scene and sparsely distribute the core factors, only using 3 for the small scenes, and 16 for the Scuol scene shown on the right side.}
\label{fig:appendix-probe-distribution}
\end{figure*}

We propose distributing light field feature probes near the camera trajectory, utilizing the Farthest Point Sampling (FPS) algorithm to selectively determine $\hat{L}$ positions. The distribution of these feature probes within the scene and their corresponding content is depicted in \figref{fig:appendix-probe-distribution}. 
To enhance the visualization of the probe distribution within target scenes, we additionally present the calibration point cloud. These calibration point clouds are not used in training and are only for visualization purposes.

\section{Ablation Details.}
We conduct ablations on key components, including the point query scheme, probe distributions, period factor, and factor aggregation module. These ablations are performed on a small indoor scene and a larger scene encompassing distant and close viewpoints. 

Our observations are as follows:

\begin{itemize}

\item Core factor modeling is critical. Disabling either factor $\ddot{\bV}$ or $\ddot{\bM}$ hampers reconstruction capabilities. Factor $\ddot{\bM}$ significantly influences overall quality, while factor $\ddot{\bV}$ improves fine-grained detail reconstruction qualitatively. 


\item Selecting factors close to the target camera yields substantial improvements, particularly for the objects with multi-scale appearances, where the camera neighborhood strategy effectively simulates a mipmap representation.

\item Smaller scenes show less sensitivity to probe distribution than larger scenes. Farthest point position sampling (FPS) provides a more uniformly distributed set of probes, facilitating stable reconstruction across different areas.

\item Coordinate scaling notably improves performance, especially in scenes captured from a small angle range. For instance, in the Free dataset, where most images are captured from a single viewing direction (as depicted in the middle of \figref{fig:appendix-probe-distribution}). The scale factor enables a more efficient use of the model capabilities.

\item Factor aggregation delivers significant performance boosts than naive feature concatenation, benefiting from the order-invariant weighted fusion design.
\end{itemize}

\begin{figure*}[!htp]
    \centering
    \resizebox{\linewidth}{!}{\includegraphics{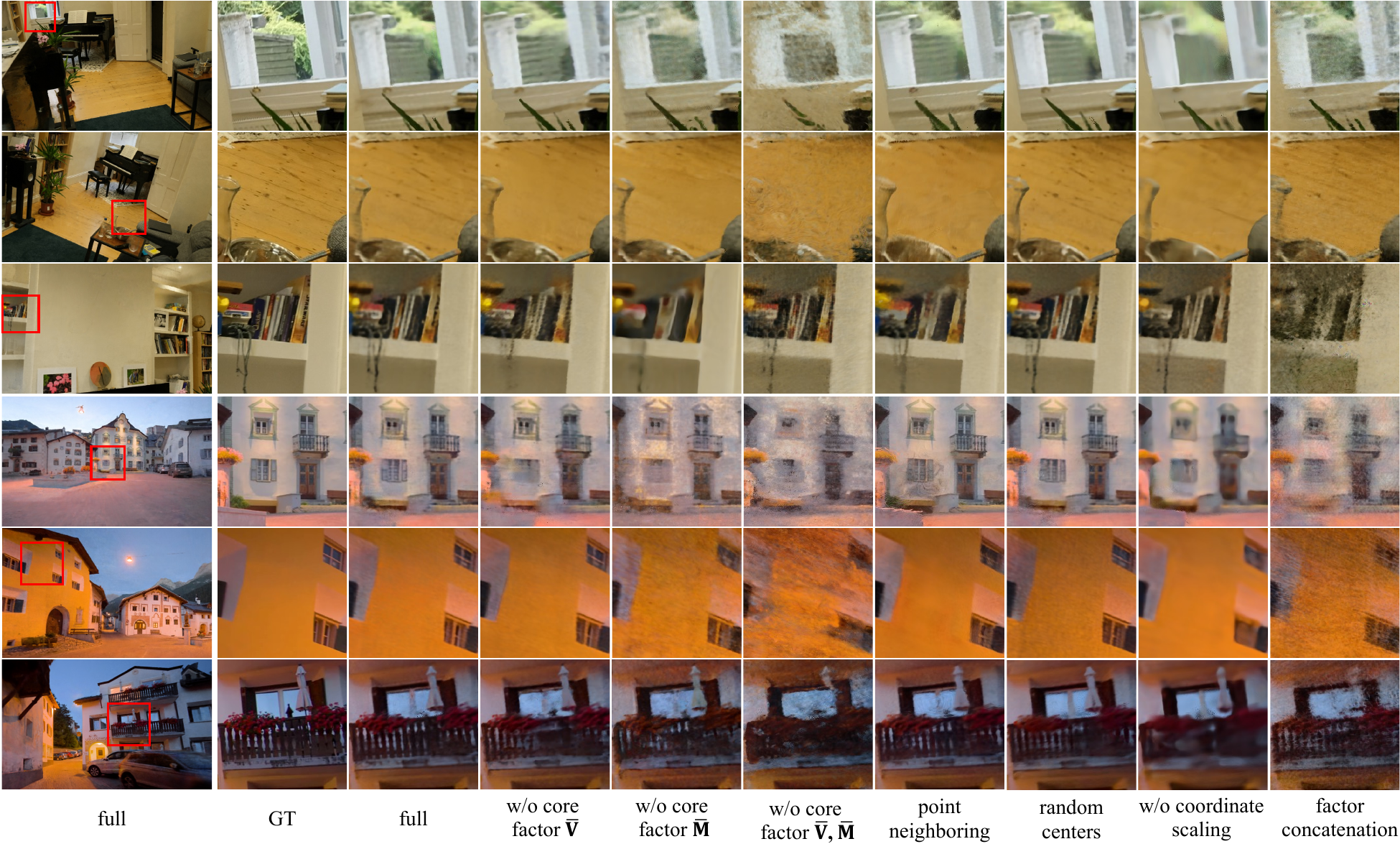}}
    \vspace{-7mm}
    \caption{\textbf{Qualitative results for ablation study.} We conduct the ablations using one indoor-small-scene and outdoor-large-scene.}
    \label{fig:appendix-qualitative-ablation}
\end{figure*}

In \figref{fig:appendix-qualitative-ablation}, we qualitatively illustrate the performance differences resulting from various design choices. Our full model provides high-quality reconstruction and better stability for diverse natural scenes that vary in extent and spatial granularity.

\section{Limitations and Future Work.}
Our \textit{NeLF-Pro} demonstrates a remarkable ability to achieve high-fidelity novel view synthesis across a wide range of natural scenes, with various levels of spatial granularity.
However, our approach currently cannot handle distractions, such as exposure changes and moving objects, and our method tends to produce `floaters' in these regions. Combining our method with techniques like feature space regularization~\cite{AlexYuandSaraFridovich-Keil2022CVPR}, appearance embedding~\cite{MartinBrualla2021CVPR} or pixel-wise loss reweighting~\cite{SabourCVPR2023} could be beneficial. In addition, our work cannot support fast rendering, it still takes about 2.6 seconds to render a $960 \times 540$ image. Incorporating fast rendering techniques, such as post-baking~\cite{Rakotosaona2023ARXIV} is an orthogonal direction to our work.

\boldstartspace{Future Work.} 
To further enhance the per-scene reconstruction quality, a possible solution is to integrate our representation with advanced multi-level modeling~\cite{Mueller2022SIGGRAPH} or cone tracing point sampler~\cite{barron2022mip360, Barron2023ICCV}. We leave this combination as future work. 
Furthermore, while this paper demonstrates success in per-scene optimization, another interesting direction for future work involves the exploration and learning of general core and basis factors across various scenes. By analyzing a large-scale dataset, we can leverage data priors not only to improve quality but also to potentially enable new applications, such as in the development of generative models.
In the ablation study, a notable finding is that matrix factors alone, without the vector core factor, can yield satisfactory reconstructions in small-scale scenes. This aspect is particularly advantageous for convolutional neural networks.

\section{More Visual Results.}
\begin{figure*}[!htp]
    \centering
    \resizebox{\linewidth}{!}{\includegraphics{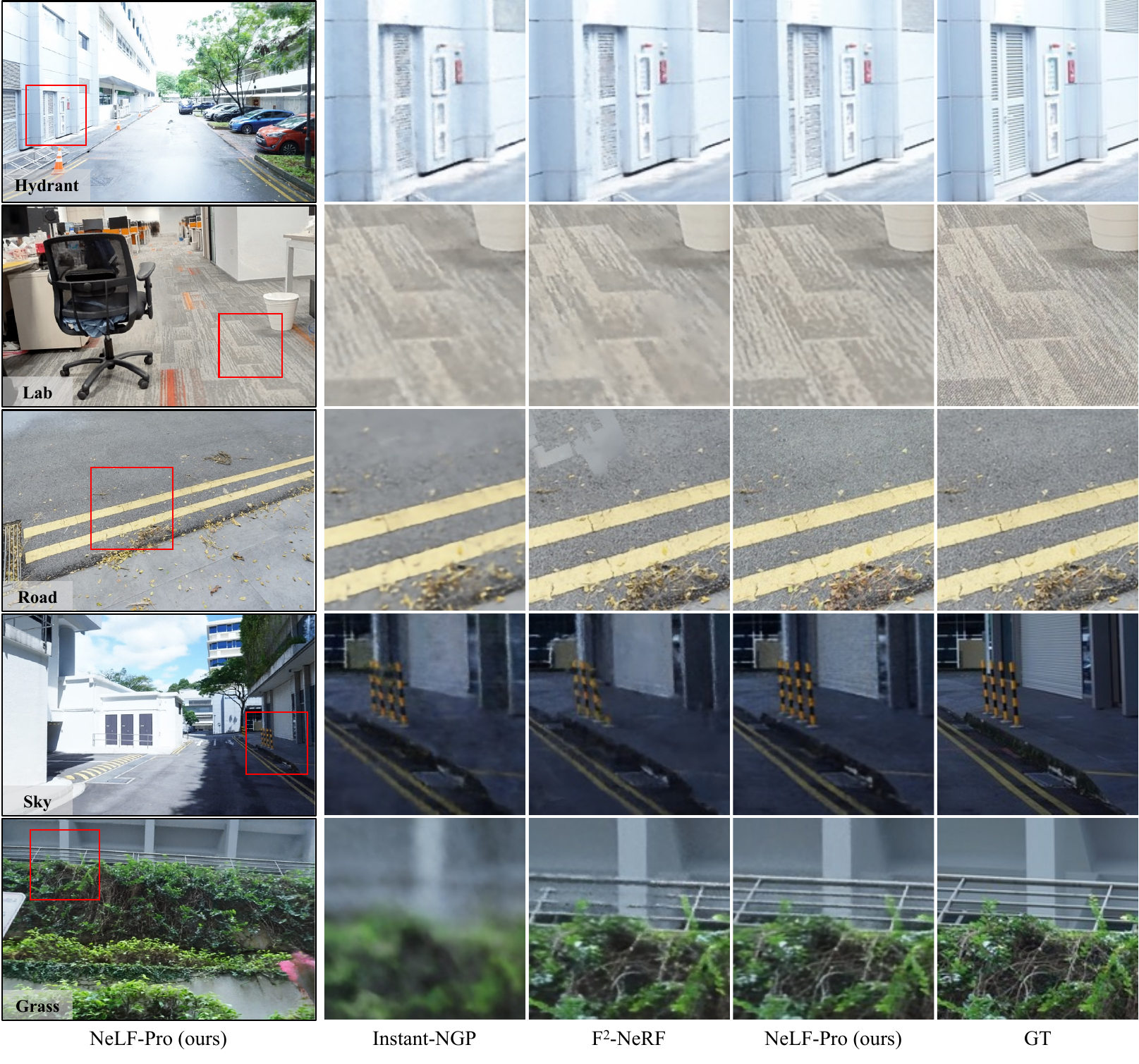}}
    \vspace{-8mm}
    \caption{\bf More Qualitative Results on the Free Dataset~\cite{Wang2023CVPR}. }
    \label{fig:appendix-qualitative-free}
\end{figure*}

\begin{figure*}[!htp]
    \centering
    \resizebox{\linewidth}{!}{\includegraphics{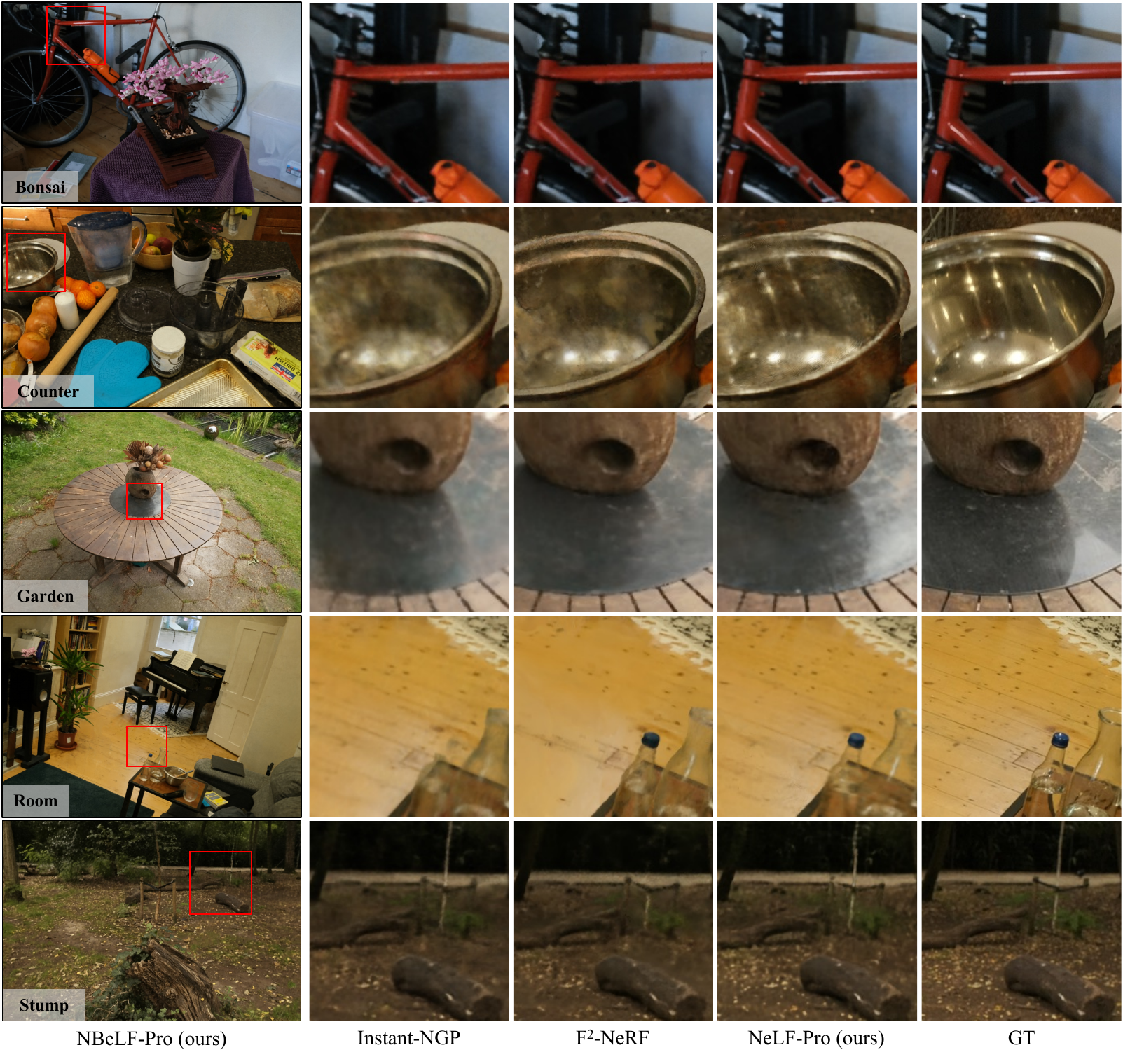}}
    \vspace{-8mm}
    \caption{\textbf{More novel view synthesis results on mip-NeRF360 Dataset~\cite{barron2022mip360}. }}
    \label{fig:appendix-qualitative-360}
\end{figure*}

\begin{figure*}[!htp]
    \centering
    \resizebox{\linewidth}{!}{\includegraphics{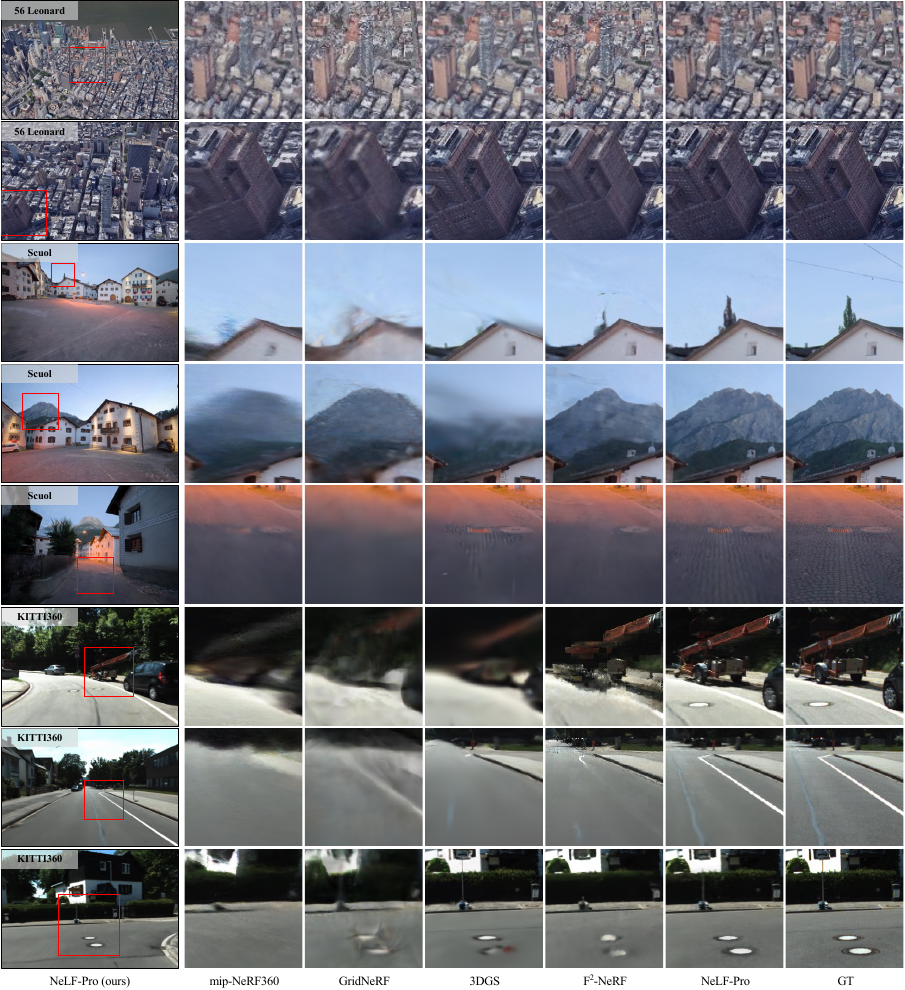}}
    \vspace{-7mm}
    \caption{\textbf{More novel view synthesis results on the Large Scale scenes.} }
    \label{fig:appendix-qualitative-large}
\end{figure*}

We show more qualitative results in \figref{fig:appendix-qualitative-free}, \figref{fig:appendix-qualitative-360}, and \figref{fig:appendix-qualitative-large}. Our \textit{NeLF-Pro} is able to better preserve detail, sharpness, and thin structures more effectively than the baselines. 

\setlength{\tabcolsep}{10pt}
\begin{table*}
    \centering
    \begin{tabular}{l|cccccccc}
    \hline
    Method & Hydrant & Lab & Pillar & Road & Sky & Stair & Grass & Avg\\
    \hline\hline
    NeRF++~\cite{Zhang2020ARXIVc} & 22.21 & 21.82 & 25.73 & 23.29 & 23.91 & 26.08 & 21.26 & 23.47\\
    mip-NeRF360~\cite{barron2022mip360} & \textbf{25.03} & \textbf{26.57} & \textbf{29.22} & \textbf{27.07} & \textbf{26.99} & \textbf{29.79} & \textbf{24.39} & \textbf{27.01}\\
    \hline
    Plenoxels~\cite{AlexYuandSaraFridovich-Keil2022CVPR} & 19.82 & 18.12 & 18.74 & 21.31 & 18.22 & 21.41 & 16.28 & 19.13\\
    DVGO~\cite{Sun2022CVPR} & 22.10 & 23.78 & 26.22 & 23.53 & 24.26 & 26.65 & 20.75 & 23.90\\
    Instant-NGP~\cite{Mueller2022SIGGRAPH}  & 22.30 & 23.21 & 25.88 & 24.24 & 25.80 & 27.79 & 21.82 & 24.43\\
    F$^2$-NeRF~\cite{Wang2023CVPR} & 24.34 & 25.92 & 28.76 & 26.76 & 26.41 & 29.19 & 22.87 & 26.32\\
    NeLF-Pro (ours) & \textbf{24.92} & \textbf{26.39} & \textbf{29.56} & \textbf{27.65} & \textbf{27.06} & \textbf{29.55} & \textbf{24.00} & \textbf{27.02}\\
    \hline
    Our SSIM & 0.770 & 0.834 & 0.818 & 0.834 & 0.873 & 0.853 & 0.629 & \textbf{0.802}\\
    Our LPIPS & 0.260 & 0.251 & 0.233 & 0.231 & 0.217 & 0.203 & 0.398 & \textbf{0.256}\\
    \hline
    \end{tabular}
    \vspace{-2mm}
    \caption[]{{\bf Per-Scene breakdown on the Free dataset.} The baseline method scores are sourced from F$^2$-NeRF; however, F$^2$-NeRF does not furnish detailed per-scene breakdowns for SSIM and LPIPS metrics.}
    \label{tab:compare_free_break_down}
    \vspace{-0em}
\end{table*}

\begin{table*}[htb]
    \centering
    \begin{tabular}{l|cccccccc}
    \hline
    Method & Bicycle & Bonsai & Counter & Garden & Kitchen & Room & Stump & Avg\\
    \hline\hline
    NeRF++~\cite{Zhang2020ARXIVc} & 22.64 & 29.15 & 26.38 & 24.32 & 27.80 & 28.87 & 24.34 & 26.21\\
    mip-NeRF360~\cite{barron2022mip360} & \textbf{23.99} & \textbf{33.06} & \textbf{29.51} & \textbf{26.10} & \textbf{32.13} & \textbf{31.53} & \textbf{26.27} & \textbf{28.94}\\
    \hline
    Plenoxels~\cite{AlexYuandSaraFridovich-Keil2022CVPR} & 21.39 & 23.65 & 25.23 & 22.71 & 24.00 &	26.38 &	20.08 & 23.35\\
    DVGO~\cite{Sun2022CVPR} & 22.12 & 27.80 & 25.76 & 24.34 & 26.00 & 28.33 & 23.59 & 25.42\\
    Instant-NGP~\cite{Mueller2022SIGGRAPH}  & 22.08 & 29.86 & 26.37 & 24.26 & 28.27 & 28.90 & 23.93 & 26.24\\
    F$^2$-NeRF~\cite{Wang2023CVPR} & 22.11 & 29.65 & 25.36 & 24.76 & 28.97 & 29.30 & \textbf{24.60} & 26.39\\
    NeLF-Pro (ours) & \textbf{22.42} & \textbf{31.28} & \textbf{27.52} & \textbf{25.08} & \textbf{29.79} & \textbf{30.10} & 24.58 & \textbf{27.27}\\
    \hline
    Our SSIM & 0.496 & 0.907 & 0.823 & 0.691 & 0.868 & 0.871 & 0.615 & \textbf{0.753} \\
    Our LPIPS & 0.500 & 0.289 & 0.366 & 0.331 & 0.240 & 0.367 & 0.427 & \textbf{0.360}\\
    \hline
    \end{tabular}
    \vspace{-2mm}
    \caption[]{{\bf Per-Scene breakdown on the mip-NeRF360 dataset.} The baseline method scores are sourced from F$^2$-NeRF; however, F$^2$-NeRF does not furnish detailed per-scene breakdowns for SSIM and LPIPS metrics.}
    \label{tab:compare_mip_nerf360_break_down}
    \vspace{-0em}
\end{table*}

\section{Per-scene Breakdown.}
In \tabref{tab:compare_free_break_down} and \tabref{tab:compare_mip_nerf360_break_down}, we provide breakdowns of the quantity metrics for the Free dataset~\cite{Wang2023CVPR} and the mip-NeRF360 dataset~\cite{barron2022mip360}. Our \textit{NeLF-Pro} achieves consistently better rendering quality compared to previous grid-based approaches. 

\end{document}